\documentclass[journal,twoside,web]{ieeecolor}
\usepackage{generic}
\usepackage{amsmath,amssymb,amsfonts}
\usepackage{algorithmic}
\usepackage{graphicx}
\usepackage{textcomp}
\usepackage{caption}
\usepackage{tabularx}
\usepackage{multirow}
\usepackage[table]{xcolor}
\def\BibTeX{{\rm B\kern-.05em{\sc i\kern-.025em b}\kern-.08em
    T\kern-.1667em\lower.7ex\hbox{E}\kern-.125emX}}

\usepackage[colorlinks=true,citecolor=blue,linkcolor=blue,urlcolor=blue]{hyperref} 
\usepackage{booktabs}
\usepackage{cuted}    
\usepackage{capt-of}  

\usepackage[backend=bibtex, style=ieee, sorting=none, citestyle=numeric-comp]{biblatex}
\definecolor{mygray}{gray}{0.95}        
\definecolor{mypink}{rgb}{0.858, 0.188, 0.478} 

\newcommand{\inc}[1]{\textcolor{mypink}{\scriptsize~+#1}}

\begin{document}
\title{VLADriver-RAG: Retrieval-Augmented Vision-Language-Action Models for Autonomous Driving}
\author{
\textbf{Rui Zhao}$^{1,2}$, \textbf{Haofeng Hu}$^{1,2}$, \textbf{Zhenhai Gao}$^{1,2}$, \textbf{Jiaqiao Liu}$^{3}$, and \textbf{Gao Fei}$^{1,2}$\\[1.2em]
$^{1}$College of Automotive Engineering, Jilin University,\\[0.15em]
$^{2}$The National Key Laboratory of Automotive Chassis Integration and Bionics, Jilin University,\\[0.15em]
$^{3}$ReeFocus AI Technology
}
\maketitle


\begin{abstract}
Vision-Language-Action (VLA) models have emerged as a promising paradigm for end-to-end autonomous driving, yet their reliance on implicit parametric knowledge limits generalization in long-tail scenarios. While Retrieval-Augmented Generation (RAG) offers a solution by accessing external expert priors, standard visual retrieval suffers from high latency and semantic ambiguity. To address these challenges, we propose \textbf{VLADriver-RAG}, a framework that grounds planning in explicit, structure-aware historical knowledge. Specifically, we abstract sensory inputs into spatiotemporal semantic graphs via a \textit{Visual-to-Scenario} mechanism, effectively filtering visual noise. To ensure retrieval relevance, we employ a \textit{Scenario-Aligned Embedding Model} that utilizes Graph-DTW metric alignment to prioritize intrinsic topological consistency over superficial visual similarity. These retrieved priors are then fused within a query-based VLA backbone to synthesize precise, disentangled trajectories. Extensive experiments on the Bench2Drive benchmark establish a new state-of-the-art, achieving a Driving Score of 89.12. The project page is available at: \url{https://github.com/InfiniDrive/VLADriver-RAG}.
\end{abstract}


\section{introduction}
\label{sec:introduction}
\IEEEPARstart{R}{ecent} strides in Vision-Language Models (VLMs) have endowed multimodal foundation models with open-world knowledge and zero-shot generalization capabilities \cite{chen2024internvl,wang2024qwen2}. Incorporating these pre-trained paradigms into autonomous driving frameworks presents a promising avenue to bolster system resilience, particularly when navigating heterogeneous traffic scenarios \cite{fu2025orion}. Moreover, this integration facilitates more intuitive human-vehicle interaction. 

Compared to traditional end-to-end autonomous driving models that mainly rely on task-specific implicit representations for perception, decision-making, and planning \cite{jia2023driveadapter, jia2023driveadapter, jia2023think, hu2023planning,jiang2023vad,zhai2023rethinking}, the paradigm is now shifting toward Vision-Language-Action (VLA) models, which incorporate explicit semantic understanding into physical control. Capitalizing on the foundational reasoning capabilities of VLMs, recent frameworks have validated that enabling models to reason in 3D space and align with behavioral planning states significantly enhances their decision-making robustness in open-world scenarios~\cite{wang2024omnidrive, fu2025orion, wang2023drivemlm}. Unlike conventional VLMs, which primarily interpret the driving environment through textual outputs, VLA models extend this capability to action generation by directly producing control commands or trajectory waypoints~\cite{black2024pi_0, intelligence2025pi_}. By coupling semantic understanding with embodied decision-making, VLA models enable a tighter integration between perception and execution, leading to more effective end-to-end autonomous driving. By unifying perception and execution equipped with semantic reasoning, VLAs serve as the next-generation end-to-end foundation models for autonomous driving~\cite{renz2025simlingo}.

\begin{figure}
    \centering
    \includegraphics[width=\linewidth]{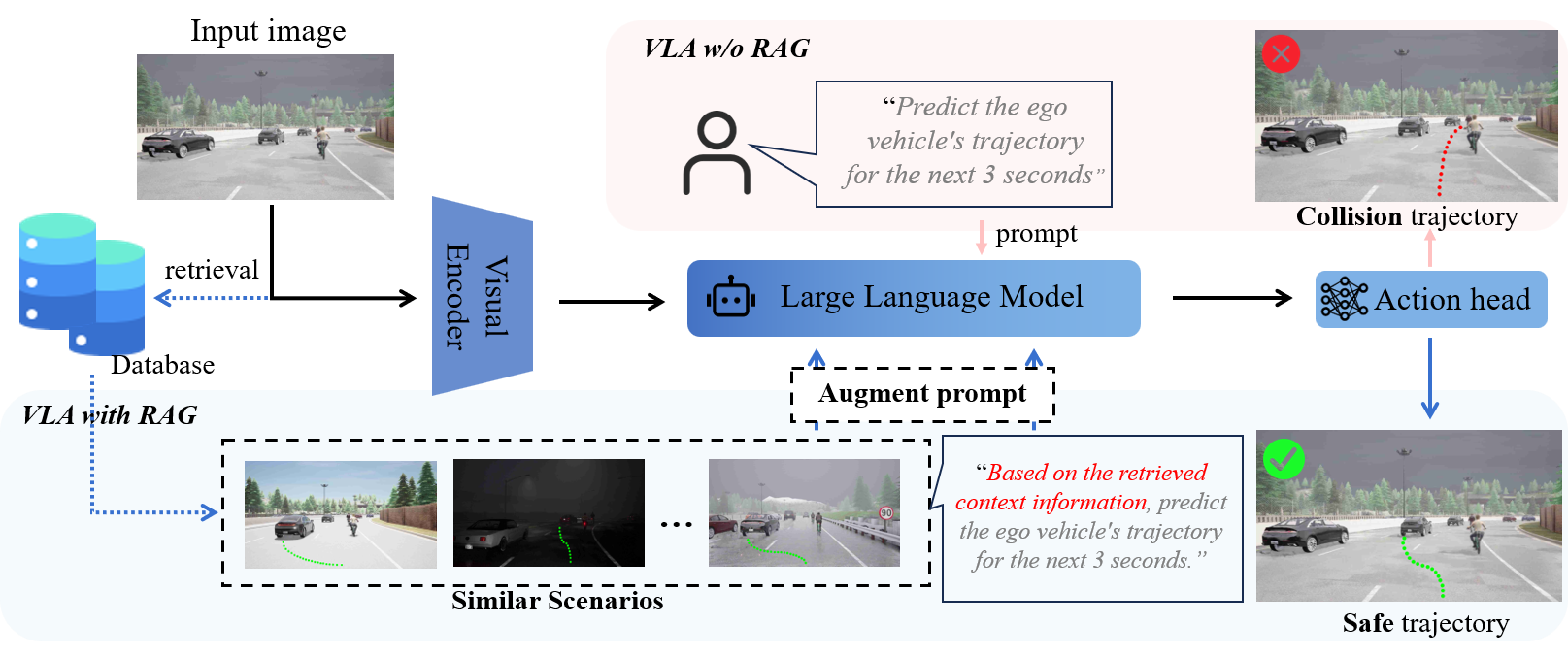}
    \caption{Conceptual comparison between a standard VLA model (Top) and our proposed RAG-augmented VLA framework (Bottom).}
    \label{fig:withoutRAG}
\end{figure}
Nevertheless, despite their advancements, both VLMs and VLA architectures face a shared critical bottleneck: the long-tail distribution of driving data \cite{jia2023think,li2024ego}. Dependence on implicit parametric knowledge often limits the model's ability to generalize to rare scenarios that are sparsely represented in training datasets. Such reliance renders the model susceptible to catastrophic forgetting and ungrounded decision-making when exposed to out-of-distribution (OOD) scenarios.

To surmount these constraints, the research community has increasingly adopted Retrieval-Augmented Generation (RAG). Initially developed for Large Language Models (LLMs) with external textual knowledge bases \cite{sun2025curriculum,peifeng2024joint}, RAG retrieves relevant evidence to ground generation, reducing hallucinations and improving factual accuracy. It also has recently shown promise in extending the capabilities of embodied agents \cite{hussien2025rag,yuan2024rag}. By accessing a non-parametric memory of past experiences, these systems can theoretically transcend the limitations of fixed training sets, effectively mitigating the challenges posed by long-tail distributions and rare corner cases. As illustrated in Fig.\ref{fig:withoutRAG}, this capability establishes a critical advantage over standard non-augmented VLA architectures. While a baseline model, constrained by parametric limitations, risks hallucinating unsafe trajectories in rare, open-world scenarios, \textit{VLA with RAG} leverages these retrieved expert priors to synthesize grounded maneuvering policies, thereby providing the foundation model with reliable, context-aware guidance during critical inference phases.  

\begin{figure*}
    \centering
    \includegraphics[width=0.8\linewidth]{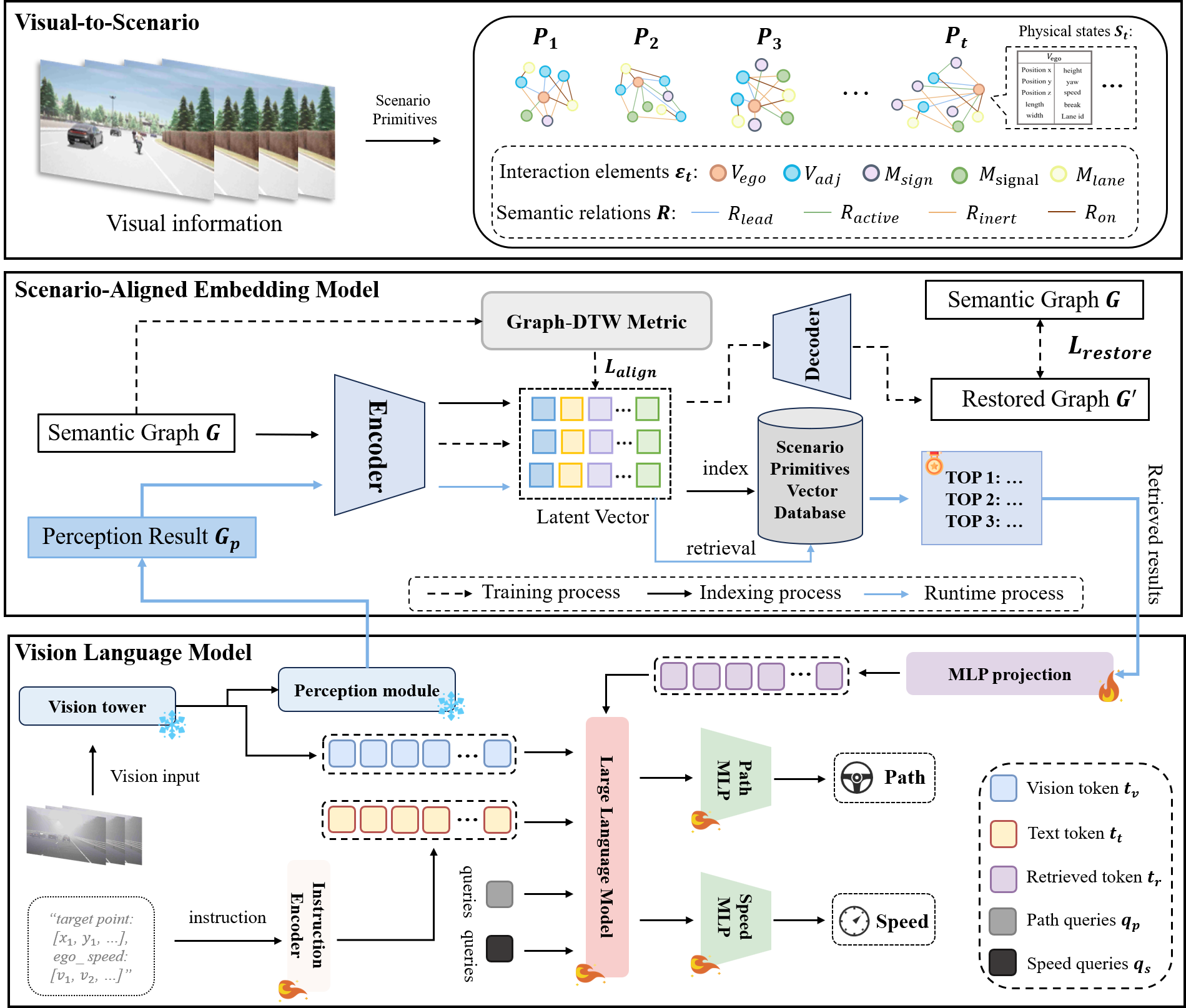}
    \caption{\textbf{The overall architecture of the proposed framework.} The system is orchestrated across three interconnected components: The \textbf{Visual-to-Scenario} module abstracts unstructured visual data into discrete spatiotemporal semantic graphs, representing scenario primitives (Top). The \textbf{Scenario-Aligned Embedding Model} projects these semantic graphs into a metric-preserving latent space supervised by a Graph-DTW metric, enabling the retrieval of pertinent historical driving priors (Middle). The \textbf{Vision Language Model} synergizes real-time vision tokens, text instructions, and retrieved context tokens, leveraging the LLM to synthesize the final path and speed planning (Bottom).}
    \label{fig:framework}
\end{figure*}

Despite the theoretical promise of retrieval-augmented frameworks, their direct adaptation to the high-stakes domain of autonomous driving is impeded by two fundamental challenges. First, regarding real-time efficiency, the high dimensionality of dense visual streams imposes excessive computational overhead, rendering raw retrieval incompatible with the strict millisecond-level latency constraints of closed-loop control~\cite{yu2024visrag, jia2025spatial}. Second, regarding scenario distinguishability, direct visual matching often suffers from high visual similarity; scenarios with identical static backgrounds but distinct semantic logic (e.g., different traffic light phases) are difficult to differentiate based solely on pixel-level features, leading to the retrieval of confused guidance~\cite{chang2025driving}.

To address these challenges, we propose VLADriver-RAG, a Retrieval-Augmented Vision-Language-Action framework specifically engineered for autonomous driving. Acknowledging that visual data constitutes the primary information carrier in driving tasks, our paradigm seeks to operationalize this dense sensory information rather than discarding it. The core philosophy of our approach is to shift the retrieval focus from surface-level visual appearance to the underlying interaction logic. By abstracting continuous observations into discrete semantic representations, we establish a retrieval mechanism governed by geometric consistency. This ensures that the system identifies historical scenarios sharing similar spatial constraints and navigation challenges. 

Our contributions are as follows:
\begin{itemize}
    \item We propose a "Visual-to-Scenario" construction mechanism that distills high-dimensional sensory streams into structured semantic snapshots. By extracting key driving elements and their spatial attributes, this mechanism bridges the semantic gap between raw pixels and decision logic, transforming visual data into a queryable format for efficient, noise-robust indexing.
\end{itemize}

\begin{itemize}
    \item We design a RAG module underpinned by a Scenario-Aligned Embedding Model. Leveraging a Relation Graph Convolutional Network (R-GCN), this model encodes semantic graphs into a compact latent space. By explicitly capturing geometric topology and interaction logic, the model establishes a discriminative metric space where scenario differentiation is determined by vector distance. This mechanism effectively disentangles visually akin but topologically distinct scenarios by enforcing a large distance margin between them, ensuring alignment based on intrinsic maneuvering context rather than superficial pixel-level similarity.
\end{itemize}
\begin{itemize}
    \item We propose a retrieval-augmented VLA architecture to bridge the gap between fixed parametric knowledge and open-world generalization. By integrating direct perception with dynamically retrieved context-aware expert demonstrations, the proposed framework enables more reliable planning in sparse and challenging corner cases. Experiments on Bench2Drive show that our method achieves 89.12 DS and 70.42\% SR, outperforming state-of-the-art (SOTA) baselines.
\end{itemize}

\section{related work}
\label{sec:introduction}
\subsection{Vision-Language-Models in Autonomous Driving}

Motivated by the exceptional contextual reasoning and comprehensive world knowledge intrinsic to VLMs, recent research has extensively integrated them into autonomous driving systems. One line of work \cite{hwang2024emma,wang2024omnidrive,xing2025openemma} directly employs VLMs for environmental perception and explainable trajectory prediction; for instance, Omnidrive \cite{wang2024omnidrive} bridges the 3D vision-language gap to generate textual trajectories via feature compression, while EMMA \cite{hwang2024emma} utilizes Gemini \cite{team2023gemini} to output discrete motion plans with strong open-loop performance. ORION \cite{fu2025orion} aligns the semantic reasoning and numerical action spaces to generate precise multi-modal trajectories via a generative planner instructed by a specialized planning token. 

To better balance high-level reasoning with low-level control, recent studies have increasingly adopted hierarchical architectures that integrate VLMs with downstream planners. Initially, this paradigm leveraged VLMs as semantic advisors that output explicit natural language guidance, such as graph-structured visual question answering in DriveLM \cite{sima2024drivelm} or interpretable textual driving intentions in DriveGPT4 \cite{xu2024drivegpt4}. Building upon this language-guided foundation, subsequent frameworks advanced to generating discrete spatial or behavioral guidance—such as coarse waypoints in DriveVLM \cite{tian2024drivevlm} or meta-actions in Senna \cite{jiang2024senna}—which are subsequently instantiated into smooth trajectories by a low-level planner. While frameworks like DriveMLM \cite{wang2023drivemlm} and LMDrive \cite{shao2024lmdrive} successfully extended these discrete strategies to closed-loop settings, these stratified methods often struggle to bridge the fundamental gap between high-level semantic tokens and high-frequency vehicular dynamics. To eliminate this cascading error, the paradigm is shifting towards continuous VLA models. Functioning as fully embodied agents, continuous VLAs directly translate multi-modal semantic understanding into execution—such as directly outputting 3D dense trajectory waypoints in OmniDrive \cite{wang2024omnidrive} or predicting continuous physical control commands in Orion \cite{fu2025orion}—thereby achieving a seamless unification of reasoning and physical action.

\subsection{Retrieval-Augment Generation in Large Language Model}
Addressing the limitations of static parametric knowledge and the hallucination issues inherent in LLMs, RAG has emerged as a dominant paradigm to incorporate external information. Foundational frameworks \cite{guu2020retrieval,lewis2020retrieval,borgeaud2022improving} typically adopt a "retrieve-then-generate" pipeline, where a dense retriever fetches relevant passages to condition the generation process. For instance, REALM \cite{guu2020retrieval} integrates a latent knowledge retriever into the pre-training stage, while the seminal RAG \cite{lewis2020retrieval} combines pre-trained parametric memory with non-parametric data. To further enhance the synergy between retrieval and generation, recent studies have evolved towards dynamic approaches: RETRO \cite{borgeaud2022improving} scales up capabilities via cross-attention chunk retrieval, FLARE \cite{jiang2023active} actively decides when to retrieve based on generation confidence, and Self-RAG \cite{asai2024self} introduces reflection tokens to critique retrieved content.

However, extending this paradigm to the demanding domain of autonomous driving presents significant challenges. Recent frameworks have attempted to bridge this gap; for example, RAG-Driver \cite{yuan2024rag} leverages retrieved expert demonstrations to enhance interpretability via in-context learning, and Spatial Retrieval AD \cite{jia2025spatial} introduces a map-based paradigm augmenting onboard sensors with offline geographic images. Despite these advancements, most existing methods struggle to simultaneously reconcile retrieval efficiency with semantic fidelity, as computational latency and topological mismatches inevitably degrade the closed-loop performance of autonomous driving systems. To surmount these bottlenecks, we propose VLADriver-RAG, a retrieval-augmented VLA architecture that distills high-dimensional visual streams into structured semantic graphs and leverages a topology-aware embedding model to retrieve geometrically consistent expert demonstrations, thereby establishing a dual-stream reasoning paradigm for robust open-world decision-making.
\begin{figure*}
    \centering
    \includegraphics[width=0.9\linewidth]{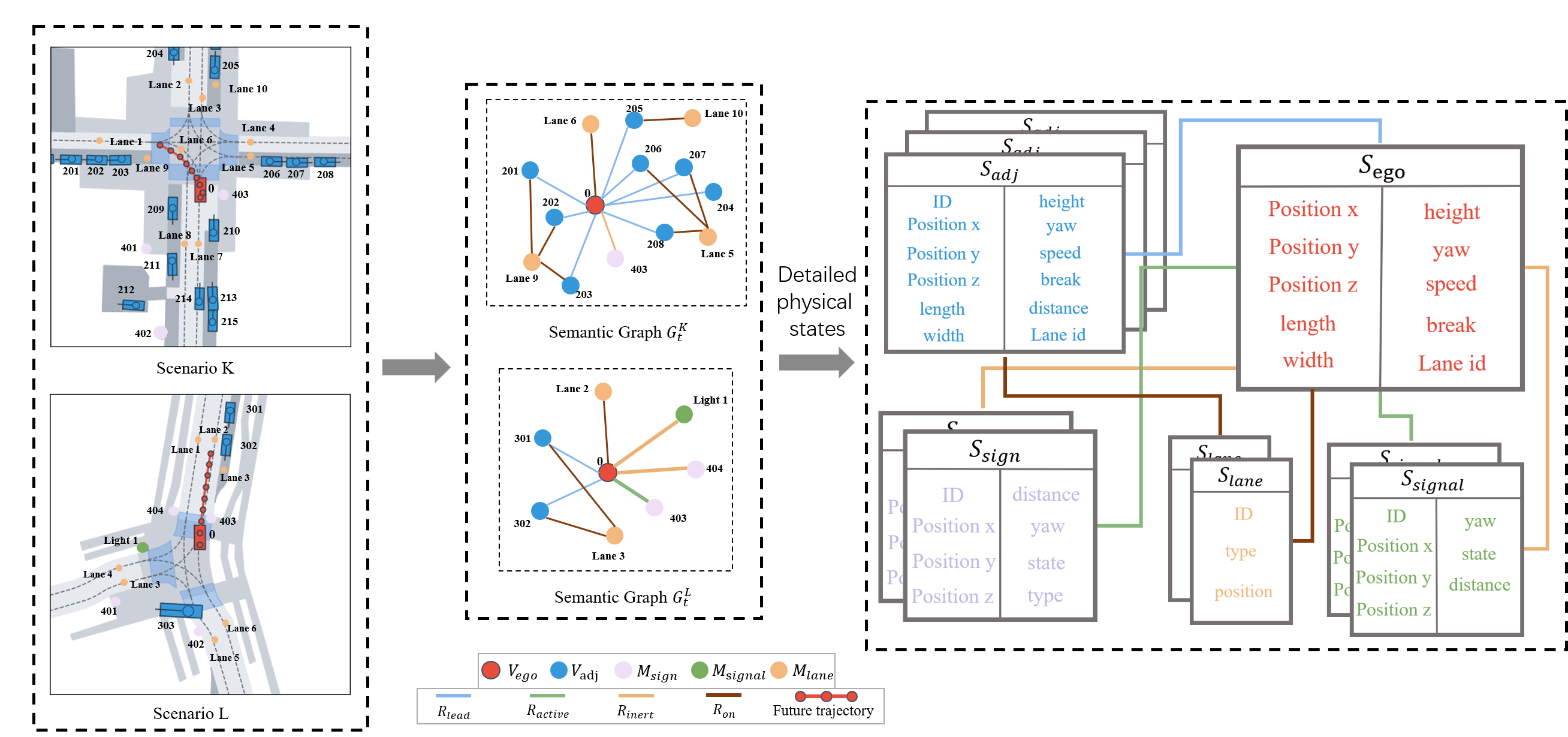}
    \caption{The detailed process of Visual-to-Scenario and Physical States}
    \label{fig3:detailed}
\end{figure*}

\section{method}      
The fundamental goal of an autonomous driving task is to synthesize a smooth, collision-free trajectory over a future horizon, conditioned on the current observation and intent.
However, exclusive reliance on implicit parametric knowledge limits generalization in long-tail scenarios. To address this, we reformulate the task as a Retrieval-Augmented Generation problem by introducing topologically aligned retrieval $K_{ret}$ from an external expert knowledge base. Let $I_{cam}$ denote the sensor inputs, $I_{nav}$ the navigation information, and $S_{ego}$ the ego state (e.g., ego speed). The end-to-end driving policy $\Phi$ typically formulates this as a direct mapping problem:

\begin{equation}
\xi_{traj} = \Phi(I_{cam},I_{nav}, S_{ego}, K_{ret})
\end{equation}
where $\xi_{traj} \in \mathbb{R}^{T\times4}$ denotes the future $T$ time step waypoints, decoupled into path coordinates $\xi_{path} \in \mathbb{R}^{T\times2}$ and target speeds $\xi_{speed} \in \mathbb{R}^{T\times2}$, effectively bridging semantic reasoning with precise kinematic control.

As systematically depicted in Fig. \ref{fig:framework}, the framework is orchestrated across three interconnected components to support this process: first, the Visual-to-Scenario Abstraction module transforms unstructured visual data into explicit spatiotemporal semantic graphs; second, the Scenario-Aligned Embedding Model projects these representations into a metric-preserving latent space to retrieve pertinent historical driving priors; and finally, the Multi-modal Vision-Language Planner synergizes real-time perception with these retrieved contexts, leveraging the inferential capabilities of LLM to synthesize the final planning outputs.
\subsection{The Definition of Visual-to-Scenario}
To capture the intrinsic semantics of driving scenes and optimize data for model ingestion, it is essential to transform raw visual sensor inputs $I_{cam}$ into structured representations enriched with topological information. To this end, we propose the Visual-to-Scenario Mechanism. As illustrated in the ``Visual-to-Scenario'' component of Fig. \ref{fig:framework}, this mechanism functions as a semantic bridge, mapping high-dimensional visual information—representing raw driving scenarios—into structured, queryable \textbf{Scenario Primitives}, thereby facilitating efficient topological retrieval.

A Scenario Primitive is defined as the fundamental unit of a driving episode, encapsulating a complete decision-making process. As shown in Fig. \ref{fig3:detailed}, for a given scene at time $t$ (Scenario K or L), we distill visual inputs into abstract topological information, instantiated as concrete Semantic Graphs (Graph $G_t^K$ or $G_t^L$). This graph construction is ego-centric (anchored at node "0"), establishing structure by linking the ego-vehicle to surrounding traffic participants (e.g., "203", "403", or "light 1") and road topologies (e.g., lane "5", lane "6") through distinct logical relationships.
\subsubsection{Mathematical Formulation}
We define a Scenario Primitive at timestamp $t$, denoted as $P_t$, as a structured tuple capturing the instantaneous decision-making context centered on the ego vehicle. $P_t$ is formalized as:
\begin{equation}
    P_t = \{\mathcal{E}_t, \mathcal{R}_t, \mathcal{S}_t\}
\end{equation}
The components are defined as follows:

\begin{itemize}
    \item \textbf{Interaction Elements ($\mathcal{E}_t$):} The set of critical objects interacting with the ego vehicle is defined as $\mathcal{E}_t \subseteq \mathcal{E} = \{V_{ego}, V_{adj}, M_{sign}, M_{signal}, M_{lane}\}$. Here, $V_{ego}$, $V_{adj}$, $M_{sign}$, $M_{signal}$ and $M_{lane}$ denote the ego vehicle, adjacent vehicles, relevant traffic signs, traffic signals, and traffic lane geometries, respectively.
    
    \item \textbf{Semantic Relationships ($\mathcal{R}_t$):} The topological constraints governing interactions is defined as $\mathcal{R}_t \subseteq \mathcal{R} = \{R_{\mathrm{lead}}, R_{\mathrm{active}}, R_{\mathrm{inert}}, R_{\mathrm{on}}\}$. Here, $R_{\mathrm{lead}}$, $R_{\mathrm{active}}$, $R_{\mathrm{inert}}$ and $R_{\mathrm{on}}$ denote longitudinal precedence, active causal constraints, passive contextual elements and the driving-on constraint, respectively.
    
    \item \textbf{Physical States ($\mathcal{S}_t$):} To comprehensively characterize the driving environment and its constituent entities, the collective state of Interaction Elements is denoted as $\mathcal{S}_t = \{S_{ego}, S_{adj}, S_{sign}, S_{signal}, S_{lane}\}$. While $S_{ego}$ represents the singular state of the ego vehicle, the remaining components are formulated as sets containing the states of multiple observed entities at the current time step. The detailed attributes for each entity type are illustrated in Fig. \ref{fig3:detailed}."
\end{itemize}

\subsubsection{Semantic Graph Construction}
To turn this abstract definition into an actionable formulation, we project $P_t$ into a \textbf{Semantic Graph} $\mathbf{G_{t}} = (N, E)$.

\begin{itemize}
    \item \textbf{Nodes ($N$):} The elements in $\mathcal{E}_t$ are instantiated as graph nodes, storing state information $\mathcal{S}_t$ as node attributes.

    \item \textbf{Edges ($E$):} The set of directed edges $E$ instantiates the topological constraints $\mathcal{R}_t$. Each edge encodes a specific interaction type, transforming the discrete node set into a heterogeneous semantic graph. 
\end{itemize}

The Scenario Primitives offers critical benefits over raw visual embeddings by decoupling interaction logic from visual information, filtering out environmental noise. Unlike implicit feature maps, the semantic graph explicitly encodes geometric constraints , enabling the retrieval system to identify scenarios that share intrinsic navigation logic despite superficial visual disparities. Furthermore, converting dense sensory streams into sparse graph representations significantly enhances data efficiency, minimizing storage redundancy while preserving the core decision-making primitives required for planning.

\begin{figure}
    \centering
    \includegraphics[width=\linewidth]{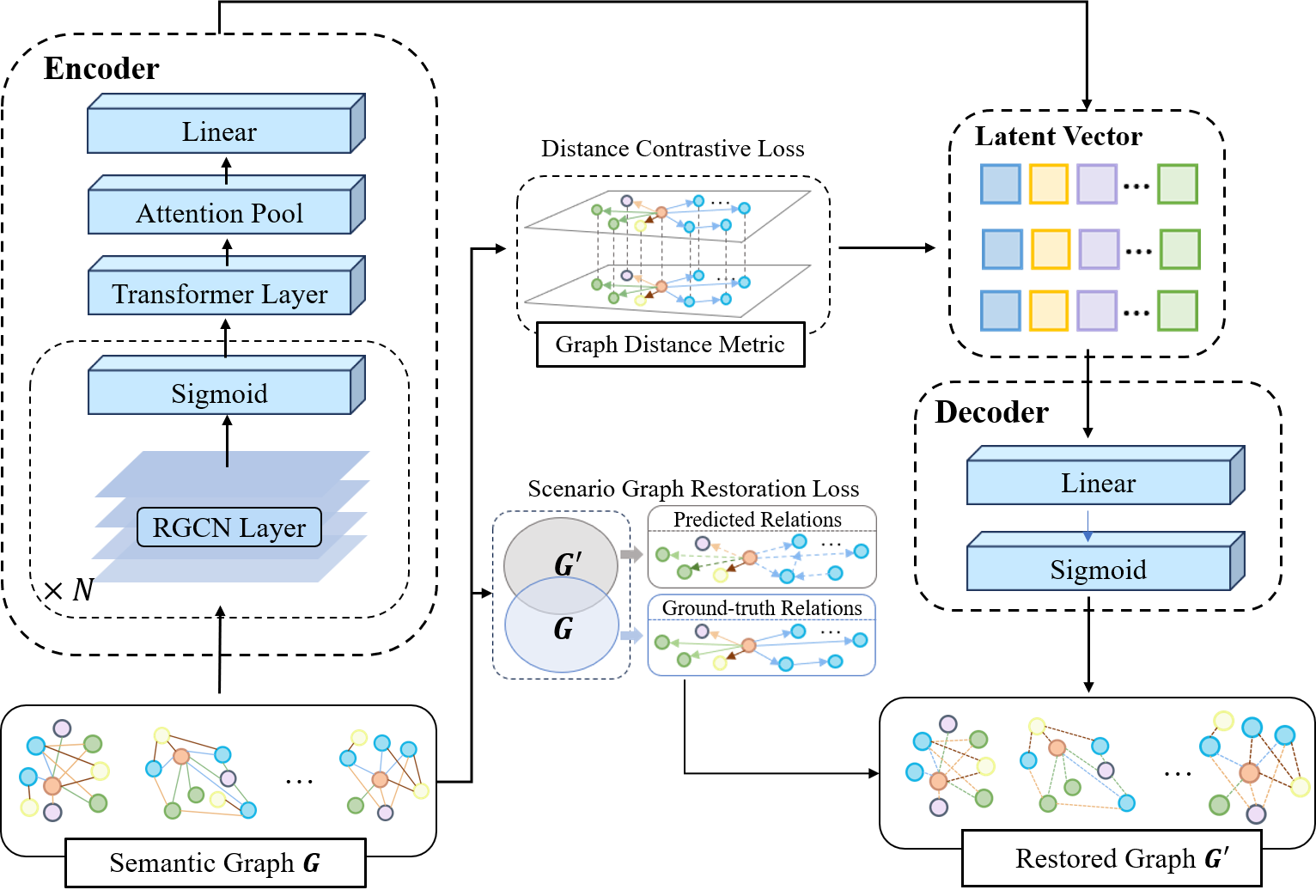}
    \caption{The detailed architecture of Embedding model}
    \label{fig:embeding_model}
\end{figure}

\subsection{Scenario-Aligned Embedding Model}
The Scenario-Aligned Embedding Model is designed to map the semantic graph $G$ into compact, measurable latent representations. Utilizing an encoder-decoder architecture, the model is trained via a dual-objective strategy that jointly optimizes self-supervised graph restoration to preserve topological semantics and Graph Dynamic Time Warping (Graph-DTW) \cite{chang2024vistascenario} metric alignment to ensure the latent space accurately reflects physical scenario similarities. In the indexing process, the trained encoder alone is employed to compress extensive historical driving priors into a searchable vector database. Subsequently, during the online runtime process, the model encodes real-time perception result $G_p$ into query vectors to retrieve relevant historical contexts, thereby augmenting the downstream LLM with planning references. The detailed architecture of the embedding model is illustrated in Fig. \ref{fig:embeding_model}.
\subsubsection{Encoder Architecture}
The primary objective of the encoder is to transform the semantic graphs into a compact latent vector that encapsulates both spatial interactions and temporal evolution. The input is defined as a sequence of semantic graphs $\{G_1, G_2, ..., G_t\}$. To capture the spatial dependencies within each frame, we employ a Relational Graph Convolutional Network (RGCN) which specifically handles the heterogeneous nature of traffic relations. For a node $i$ at layer $l+1$, the hidden feature $f_i^{l+1}$ is aggregated from its neighbors $N_i^r$ under relation $r\in \mathcal{R}$ as follows:
\begin{equation}
    f_{i}^{l+1} = \sigma \left( \sum_{r \in R} \sum_{j \in N_{i}^{r}} \frac{1}{c_{i}^{r} d_{i,j}} W_{r}^{l} f_{j}^{l} + W_{0}^{l} f_{i}^{l} \right), c_{i}^{r} = |N_i^r| 
\end{equation}
where $\sigma$ denotes the activation function, $W_r^l$ is the learnable weight matrix for relation $r$, $c_{i}^{r}$ is a normalization constant, and $d_{i,j}$ represents the interaction strength between nodes.

To capture the temporal evolution across the graph sequence, the spatially encoded features are subsequently processed by a Transformer encoder equipped with a multi-head attention mechanism with $h$ heads. This allows the model to weigh the significance of different time steps dynamically. For the $m$-th attention head, the Query, Key, and Value matrices are computed via linear projections:
\begin{equation}
    Q_{m} = q W_{m}^{q}, \quad K_{m} = k W_{m}^{k}, \quad V_{m} = v W_{m}^{v}
\end{equation}
The outputs from heads are concatenated to form the final spatiotemporal latent representation $S$:
\begin{equation}
    head_m = \text{Attention}(Q_m, K_m, V_m) = \text{softmax}\left(\frac{Q_mK_m^T}{\sqrt{d_k}}\right)V_m
\end{equation}
\begin{equation}
    S = \text{Concat}(\text{head}_1, ..., \text{head}_h) W^0
\end{equation}
where $W^0 \in \mathbb{R}^{d \times d}$ serves as a linear projection matrix that integrates the distinct contextual features captured by individual attention heads and the latent vector $S$ is the compressed representation of the scenario primitive.
\subsubsection{Self-Supervised Restoration}
To ensure that the compressed latent vector $S$ retains the fundamental semantic topology of the original scenario, we introduce a decoder module that operates in a self-supervised manner. The decoder aims to reconstruct the semantic graph $G'$ from the latent vector $S$, effectively functioning as an auto-encoder.

This reconstruction process is critical for preventing information loss during the dimensionality reduction in the encoder. The optimization objective here is to maximize the structural similarity between the input graph adjacency matrix and the predicted one.
\subsubsection{Metric-Aligned Optimization}
While the restoration loss ensures semantic fidelity, it does not guarantee that the geometric distance between two latent vectors reflects the actual physical similarity of the driving scenarios. To enable accurate retrieval, the latent space must be metric-preserving. We achieve this by aligning the latent space with Graph-DTW, a robust metric for measuring the distance between spatiotemporal graphs.

This dual-optimization strategy ensures that the resulting embeddings are both structurally informative and mathematically searchable, enabling the efficient retrieval of relevant historical driving primitives during the runtime phase.
\subsection{VLA Backbone}
\begin{figure}
    \centering
    \includegraphics[width=\linewidth]{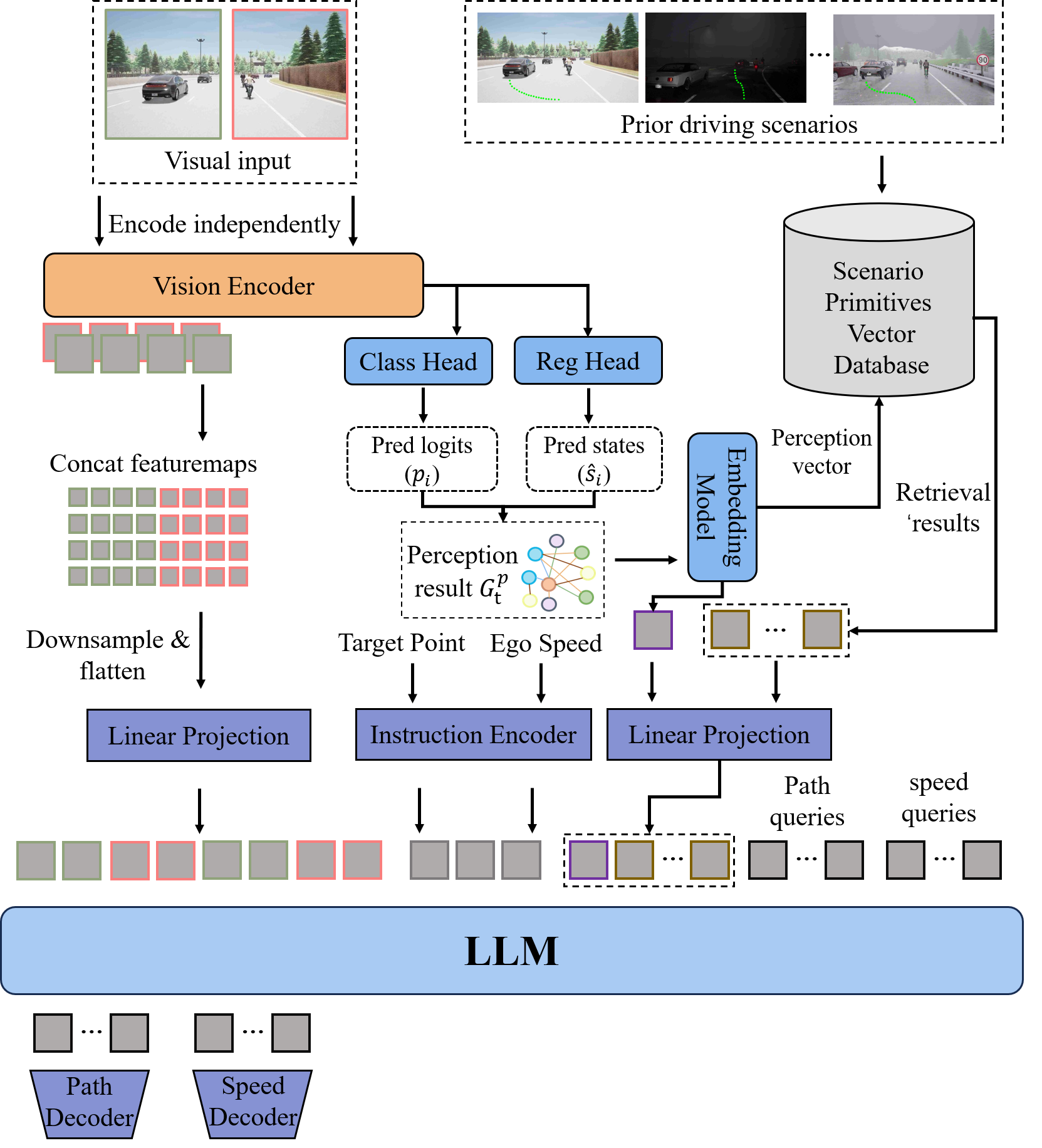}
    \caption{Overview of the VLA Backbone architecture}
    \label{fig:LLM}
\end{figure}

The VLA Backbone serves as the central reasoning engine of our framework, designed to orchestrate the transition from multi-modal perception to executable driving actions. As depicted in Fig. \ref{fig:LLM}, this module integrates visual observations, retrieved historical priors, and navigation commands into a unified embedding space. By leveraging a query-based attention mechanism, the backbone selectively aggregates salient features from this fused context, effectively distilling complex semantics into disentangled path and speed waypoints.
\subsubsection{Multi-modal Tokenization and Fusion}
To enable the LLM to reason over heterogeneous data sources, we first align visual observations, instructions, and retrieved historical priors into a unified high-dimensional embedding space. As illustrated in Fig. \ref{fig:LLM}, the input stage processes three distinct data streams via specific encoders and projection layers.

\textbf{Visual Encoding :} The raw visual input $I_{cam} \in \mathbb{R}^{H \times W \times 3}$ is processed by a frozen Vision Tower to extract high-dimensional semantic features. These features are projected directly to form the sequence of visual tokens $\mathbf{e_v} \in \mathbb{R}^{N_v \times D}$, serving as the primary visual context for the LLM.

\textbf{Perception-Aided Retrieval Encoding:} Parallel to visual token generation, the vision tower features are fed into a Perception Module with two lightweight decoder heads, as illustrated in Fig. \ref{fig:LLM}. Specifically, a classification head predicts category scores, while a regression head estimates 4-dimensional box parameters, together producing structured perception results. 

This structured graph $G_t^p$ is utilized to query the offline scenario database. The retrieved historical scenario primitives are then mapped via an MLP projection layer to form the retrieved tokens $\mathbf{e_r} \in \mathbb{R}^{N_r \times D}$, providing the model with relevant driving priors.

\textbf{Instruction Encoding:} To explicitly guide the trajectory generation, we utilize numerical navigation commands rather than natural language. The target points and ego-velocity are encoded into high-dimensional tokens using specific trainable MLPs. Formally, these kinematic inputs are projected to form the unified instruction tokens $\mathbf{e_i} = [\mathbf{e_{nav}}, \mathbf{e_{spd}}]$, where $\mathbf{e_{i}} \in \mathbb{R}^{N_I \times D}$ aligns with the embedding dimension of the backbone model.

\textbf{Fusion:} Finally, these three distinct token sequences are concatenated to construct the comprehensive input sequence $\mathbf{X_{in}}$ for the reasoning backbone:
\begin{equation}
    \mathbf{X_{in}} = [\mathbf{e_v}, \mathbf{e_i}, \mathbf{e_r}]
\end{equation}
Through dimensional alignment, the embeddings $\mathbf{e_v}$, $\mathbf{e_i}$, and $\mathbf{e_r}$ are fused into a unified sequence, which serves as the prompt for the reasoning backbone.
\subsubsection{Query-Based Reasoning Backbone}
The core of our planning framework is a generative LLM that functions as a multi-modal reasoner. Unlike standard VLM approaches that rely solely on auto-regressive text generation, we adopt a query-based mechanism to extract structured planning information efficiently.

We initialize two sets of learnable query embeddings: Path Queries $q_p \in \mathbb{R}^{N_p \times D}$ and Speed Queries $q_s \in \mathbb{R}^{N_s \times D}$. These queries are appended to the input sequence $T_{in}$ and fed into the LLM. Through the self-attention mechanism, these queries interact with the multi-modal context (vision, text, and retrieval) to aggregate specific features required for trajectory planning and velocity control, respectively.

The LLM outputs a sequence of hidden states, where the states corresponding to the query positions contain the distilled planning information:
\begin{equation}
    [H_{path}, H_{speed}] = \text{LLM}([\mathbf{X_{in}}, q_p, q_s])
\end{equation}
Here, $H_{path}$ and $H_{speed}$ represent the high-level semantic features for the vehicle's path and speed waypoints.
\subsubsection{Task-Specific Decoding and Optimization}
To translate high-level semantic features into executable control commands, we employ a disentangled output representation. The framework utilizes two lightweight MLP decoding heads to regress specific action parameters from the query outputs.
\begin{itemize}
    \item Path Head (Geometric): The path states $H_{path}$ are projected to predict future path waypoints $\xi_{path} = \{p_1, ..., p_{N_p}\}$ in the bird's-eye view (BEV) space, focusing solely on the spatial geometry of the maneuver (e.g., lane changes, turns).
    \item Speed Head (Temporal): The speed states $H_{speed}$ are projected to predict the temporal velocity waypoints $\xi_{speed} = \{v_1, ..., v_{N_s}\}$, determining how fast the vehicle traverses the path.
\end{itemize}

\subsection{Training Strategy}
To mitigate gradient instability caused by jointly optimizing multiple heterogeneous modules, we adopt a three-stage curriculum that progressively trains the embedding, perception, and planning components. In the following, we detail the objective and the role of each stage.

\textbf{Stage 1: Embedding Pre-training.}
We first train the \textit{Scenario-Aligned Embedding Model} by minimizing a composite loss
\begin{equation}
    \mathcal{L}_{total} = \lambda_r\mathcal{L}_{restore} + \lambda_a\mathcal{L}_{align},
\end{equation}
where \(\lambda_r\) and \(\lambda_a\) are weighting coefficients. The restoration loss \(\mathcal{L}_{restore}\) is built upon an Intersection over Union (IoU) objective that forces the model to preserve the dynamic interaction graph between traffic agents. Concretely, it is defined as

\begin{table}[t]
    \centering
    \caption{Closed-loop results on B2D. DS denotes Driving Score, and SR denotes Success Rate (\%).}
    \label{tab:comparison}
    
    \resizebox{\linewidth}{!}{ 
        \begin{tabular}{ll|cc}
            \toprule
            Method &  Paradigm & DS$\uparrow$ & SR$\uparrow$ \\
            \midrule
            
            AD-MLP \cite{zhai2023rethinking}  & E2E & 18.05 & 00.00 \\
            
            \rowcolor{mygray} 
            TCP \cite{wu2022trajectory}  & E2E & 40.70 & 15.00 \\
            
            TCP-traj \cite{wu2022trajectory}  & E2E & 59.90 & 30.00 \\
            
            \rowcolor{mygray}
            UniAD-Base \cite{hu2023planning}  & E2E & 45.81 & 16.36 \\
            
            VAD \cite{jiang2023vad}  & E2E & 42.35 & 15.00 \\
            
            \rowcolor{mygray}
            ThinkTwice \cite{jia2023think}  & E2E & 62.44 & 31.23 \\
            
            DriverAdapter \cite{jia2023driveadapter}  & E2E & 64.22 & 33.08 \\
            
            \rowcolor{mygray}
            ORION \cite{fu2025orion} & VLA & 77.74 & 54.62 \\

            Simlingo \cite{renz2025simlingo}  & VLA & 85.07 & 67.27 \\
            
            \midrule

            VLADriver-RAG  & VLA & \textbf{89.12}\inc{4.05} & \textbf{70.42}\inc{3.15} \\
            
            \bottomrule
        \end{tabular}
    } 
\end{table}

\begin{equation}
    \mathcal{L}_{restore} = \frac{1}{B} \sum_{k=1}^{B} \left( 1 - \frac{1}{N_{k}} \sum_{j=1}^{N_{k}} \frac{|E_{kj}^{pred} \cap E_{kj}^{gt}|}{|E_{kj}^{pred} \cup E_{kj}^{gt}|} \right),
\end{equation}
where \(B\) is the batch size, \(N_k\) denotes the number of frames in the \(k\)-th scenario, \(j\) indexes the frames within that scenario, and \(E_{kj}^{pred}\) and \(E_{kj}^{gt}\) are the predicted and ground-truth edge sets, respectively. By minimizing \(\mathcal{L}_{restore}\), the latent embedding is driven to encode enough information to reconstruct the scenario’s evolving interaction graph.

To enforce cross-scenario alignment, we introduce a contrastive distance loss that constrains the Euclidean distance between latent representations to match their Graph-DTW dissimilarity. Formally, the alignment loss is
\begin{equation}
    \mathcal{L}_{align} = \frac{2}{B(B-1)} \sum_{1 \le k < l \le B} \left( \|S_k - S_l\|_2 - d_{kl}^{DTW} \right)^2,
\end{equation}
where \(S_k\) and \(S_l\) are the latent vectors of scenario \(k\) and \(l\), and \(d_{kl}^{DTW}\) is the pre-computed Graph-DTW distance. This term ensures that structurally similar scenarios stay close in the embedding space.

\textbf{Stage 2: Perception Alignment.} With With the embedding model frozen, we shift focus to the \textit{Perception Module}. The multi-task objective for this stage is
\begin{equation}
    \mathcal{L}_{perc} = \lambda_{c} \mathcal{L}_{cls} + \lambda_{e} \mathcal{L}_{reg},
\end{equation}
where \(\lambda_{c}\) and \(\lambda_{e}\) are balancing coefficients.
The classification component \(\mathcal{L}_{cls}\) employs the Focal Loss to handle the class imbalance in interaction detection:
\begin{align}
    \mathcal{L}_{cls} &= -\frac{1}{N_{pos}} \sum_{i} \bigg[ \alpha (1-p_i)^\gamma \log(p_i) \notag \\
    &\quad + (1-\alpha) p_i^\gamma \log(1-p_i) \bigg].
\end{align}
\begin{figure}[t]
    \centering
    \includegraphics[width=\linewidth]{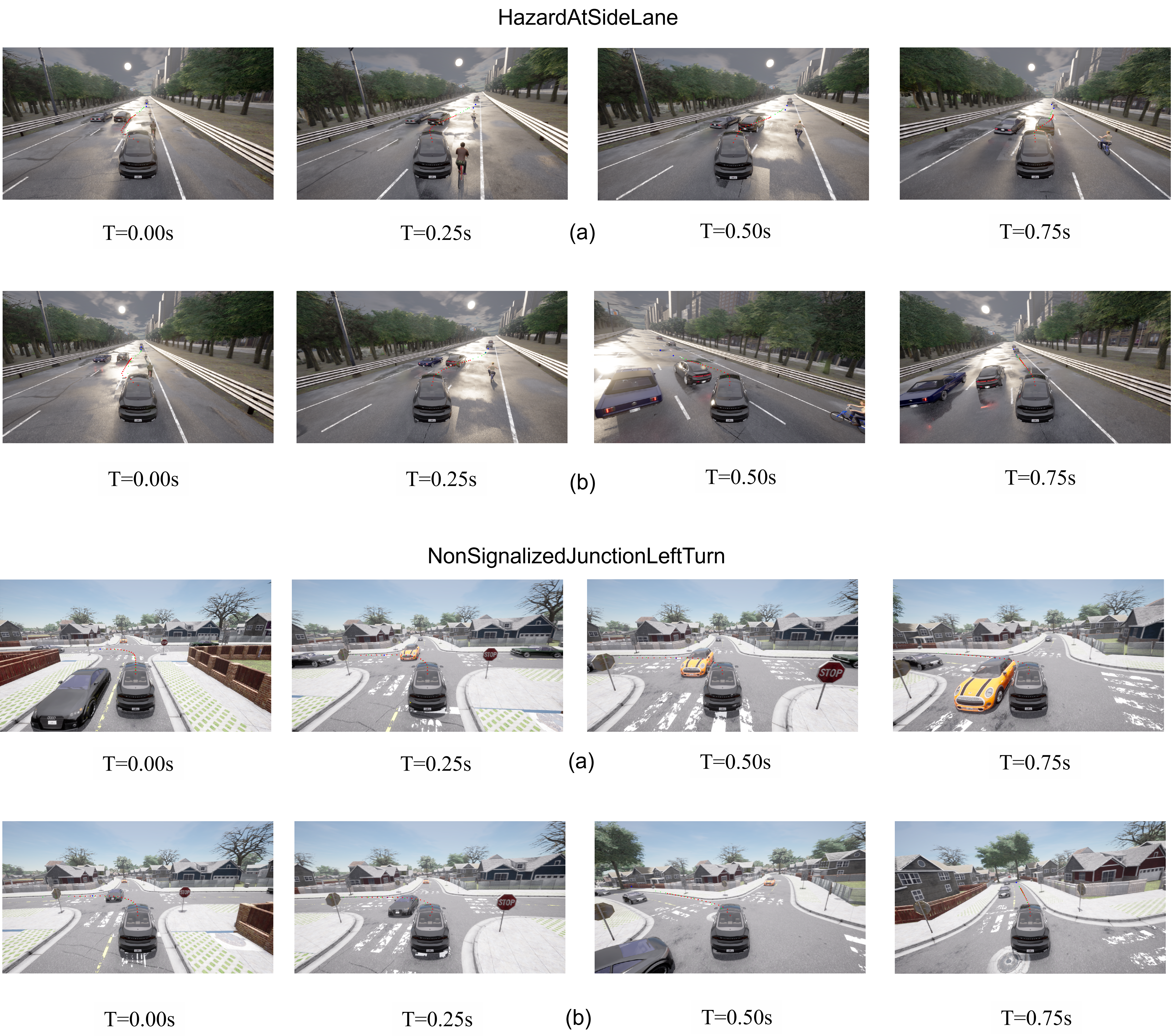} 
    \caption{\textbf{Qualitative comparison of planning trajectories in two driving scenarios.} (a) denotes the model \textit{w/o} Retrieval; (b) denotes VLADriver-RAG.}
    \label{fig:traj_vis}
\end{figure}
Here \(p_i\) is the predicted probability, \(\gamma\) is the focusing parameter, \(\alpha\) balances positive and negative samples, and \(N_{pos}\) is the total number of positive samples.
For the physical attributes, we use the SmoothL1 loss \(\mathcal{L}_{reg}\) computed over positive samples:
\begin{equation}
    \mathcal{L}_{reg} = \frac{1}{N_{pos}} \sum_{i \in \mathcal{P}} \text{SmoothL1}(\hat{\mathbf{s}}_i - \mathbf{s}_i),
\end{equation}
where \(\hat{\mathbf{s}}_i\) and \(\mathbf{s}_i\) are the predicted and ground-truth state vectors, and \(\mathcal{P}\) is the index set of positive samples. 

\textbf{Stage 3: Policy Optimization.} In the final stage, we train the model end-to-end to learn the planning policy. We adopt L2 loss for trajectory regression to improve stability against outliers, and define the total loss as a weighted combination of path and speed regression:
\begin{equation}
\mathcal{L}_{total} = \lambda_{p} \sum_{g=1}^{N_p} \| p_g - \hat{p}_g \|_2^2 + \lambda_{s} \sum_{g=1}^{N_k} \| v_g - \hat{v}_g \|_2^2,
\end{equation}
with \(\hat{p}_g,\hat{v}_g\) the ground-truth path and speed values. This disentangled supervision ensures precise lateral control via the path head and smooth longitudinal control via the speed head.

\section{Experiments}
In this section, we conduct a comprehensive evaluation to verify the effectiveness and robustness of our proposed framework in closed-loop driving scenarios. We begin by introducing the datasets and metrics. Subsequently, we provide the specific implementation details of our model architecture and training strategies. We then present a comparative analysis against SOTA methods to demonstrate our superior planning performance. Finally, we conduct ablation studies to investigate the contribution of each core component.
\label{sec:experiments}

\begin{table}[h]
    \centering
    \caption{\textbf{Ablation studies on the VLADriver-RAG system.} GBR: employing the proposed Graph-Based Retrieval which matches topological structures. VSR: utilizing Visual-Similarity Retrieval based on raw image feature cosine similarity. Emb-Rec: training the embedding model using exclusively the restoration loss. Emb-Full: training the embedding model with the total composite loss objectives.}
    \label{tab:abla1}
    \resizebox{\linewidth}{!}{ 
    \begin{tabular}{c|cccc|cc}
        \toprule
        & GBR & VSR & Emb-Rec & Emb-Full & DS$\uparrow$ & SR$\uparrow$ \\ 
        \midrule
      \textit{w/o} Retrieval  & -  & -  & -  &  - & 85.14  &  66.56 \\ 
        \cmidrule{1-7} 
        \multirow{3}{*}{with Retrieval} & -  & \checkmark  & -  &  \checkmark &  86.65 & 67.21  \\ 
                          & \checkmark  & -  & \checkmark  & -  & 88.92  & 69.10  \\ 
        &  \checkmark &  - &  - &  \checkmark & \textbf{89.12}  &  \textbf{70.42} \\ 
        \bottomrule
    \end{tabular}
    }
\end{table}

\subsection{Datasets and Metrics}
We evaluate our approach exclusively on the closed-loop simulation benchmark Bench2Drive (B2D), a CARLA-based platform designed for rigorous end-to-end driving assessment. The dataset comprises 10,000 sampled driving segments for training, while the evaluation set consists of 220 short routes, each characterizing a distinct and safety-critical driving scenario that challenges the agent's planning capabilities. To quantitatively measure performance, we adopt the two official metrics defined by the benchmark: Success Rate (SR), which reflects the percentage of driving tasks successfully completed without collision or critical failure, and Driving Score (DS), a composite measure that integrates route completion progress with penalties for driving infractions, providing a granular assessment of both safety and efficiency.

\subsection{Implementation Details}
We implement our framework using PyTorch. For the visual component, we utilize CLIP-ViT as the vision encoder. The reasoning backbone is a lightweight 50M-parameter Transformer decoder based on the LLaMA architecture, which is initialized and trained from scratch. The comprehensive input sequence $\mathbf{X_{in}}$ is injected into this backbone. 

We follow the progressive training strategy described in Sec. III-D. 
Specifically, the embedding model (Stage 1) is pre-trained for 10 epochs with a batch size of 128. 
Subsequently, the perception module (Stage 2) is optimized while keeping the vision encoder frozen. 
Finally, for the end-to-end policy learning (Stage 3), we train for 30 epochs with a global batch size of 48. 
We use the AdamW optimizer with an initial learning rate of $3e^{-5}$.

For the optimization objective, we balance the regression tasks by setting the loss weights $\{\lambda_r,\lambda_a,\lambda_c,\lambda_e,\lambda_p,\lambda_s\}=\{1,1,1,5,1,1\}$. In the query-based reasoning module, we initialize $N_p = 20$ path queries and $N_s = 10$ speed queries to cover the planning horizon. For the retrieval-augmented component, we construct a scenario database containing 800,000 primitives, from which the top $N_r = 10$ most relevant priors are retrieved for each query. The multi-head attention mechanism operates with $h=8$ heads. All experiments are conducted on a server equipped with $2\times$ NVIDIA A100 GPUs.

\subsection{Main Results on Driving Benchmarks}
As quantitatively summarized in Table \ref{tab:comparison}, our framework establishes a new SOTA on the B2D benchmark, surpassing the strong VLA baseline Simlingo \cite{renz2025simlingo}. Specifically, we achieve a SR of 70.42\% and a DS of 89.12, representing significant gains of 3.15\% and 4.05, respectively.

The comprehensive improvement in both safety and driving quality is likely related to the proposed graph-based retrieval mechanism. By retrieving and conditioning on relevant historical scenarios, the model acquires high-quality, structure-aware priors that serve as explicit references. This external knowledge not only helps the planner resolve perceptual ambiguities in long-tail scenarios—thereby preventing failures and boosting SR—but also guides the generation of more compliant and human-like trajectories, which directly translates to a higher DS by minimizing infractions and optimizing route progress.

\subsection{Ablation}
To rigorously validate the architectural design and optimization strategies of our framework, we conduct extensive ablation studies on the B2D benchmark. We structure our analysis into three dimensions: the design of the retrieval mechanism, the objective of knowledge embedding, and the training curriculum.

\subsubsection{design of the retrieval mechanism}

We perform a systematic decomposition of our framework, as detailed in Table \ref{tab:abla1}, to isolate and verify the contribution of each component. This analysis specifically scrutinizes the impact of the retrieval mechanism itself, the retrieval modality efficiency, and the embedding optimization objectives.

\paragraph{Impact of Retrieval-Augmented Paradigm.}
We first evaluate the effect of the retrieval mechanism by comparing the baseline with the full model. As shown in Table \ref{tab:abla1}, the \textit{w/o Retrieval} variant (row 1), which relies only on current visual inputs, achieves the worst performance, while VLADriver-RAG (row 4) delivers clear improvements. This result indicates that historical references are crucial for robust decision-making in complex long-tail scenarios. The qualitative examples in Fig. \ref{fig:traj_vis} further support this finding: the baseline fails in corner-case driving scenarios, whereas VLADriver-RAG successfully plans a safe trajectory. Together, these results show that retrieved knowledge effectively improves planning stability in uncertain environments.

\paragraph{Retrieval Modality Efficiency.}
Next, we investigate the retrieval modality by comparing the Visual-Similarity Retrieval (VSR) variant (row 2) in against our proposed Graph-Based Retrieval (GBR) employed in the full model (row 4). 
Specifically, in the VSR setting, we construct a visual features database with the same size as in the previous experiments, and use CLIP-ViT as the embedding model. Retrieval is then performed by computing and ranking the cosine similarity between the current observation embedding and the stored visual features.
Results indicate that relying purely on visual similarity yields suboptimal gains. We attribute this to the fact that visually similar scenes may require vastly different driving logic. In contrast, our GBR approach ensures that priors are topologically consistent with the current challenge.

\begin{figure}[t]
    \centering
    \includegraphics[width=0.8\linewidth]{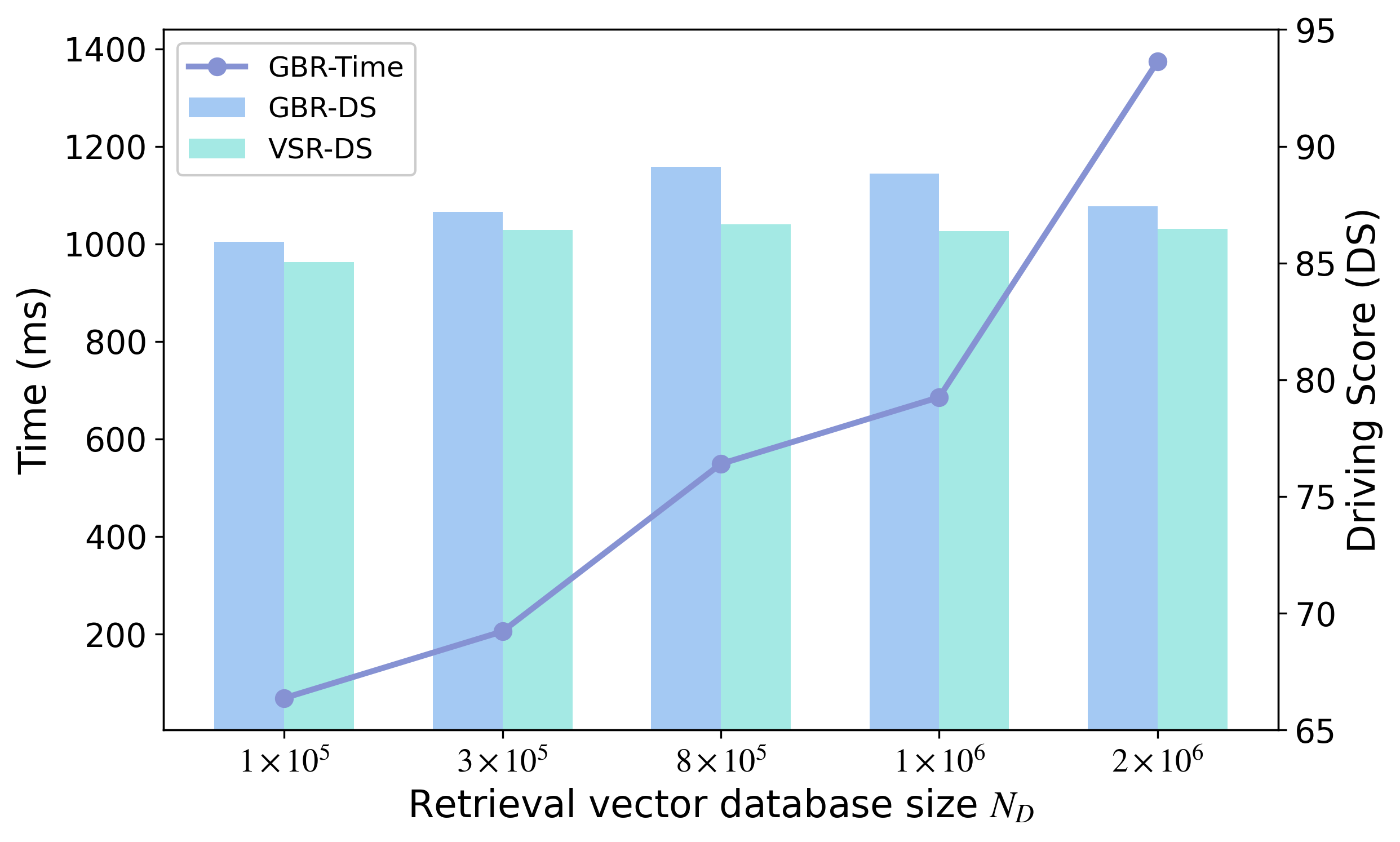} 
    \caption{Comparison of retrieval time and driving score under different retrieval modality}
    \label{fig:speed_comp}
\end{figure}
\begin{figure}[t]
    \centering
    \includegraphics[width=\linewidth]{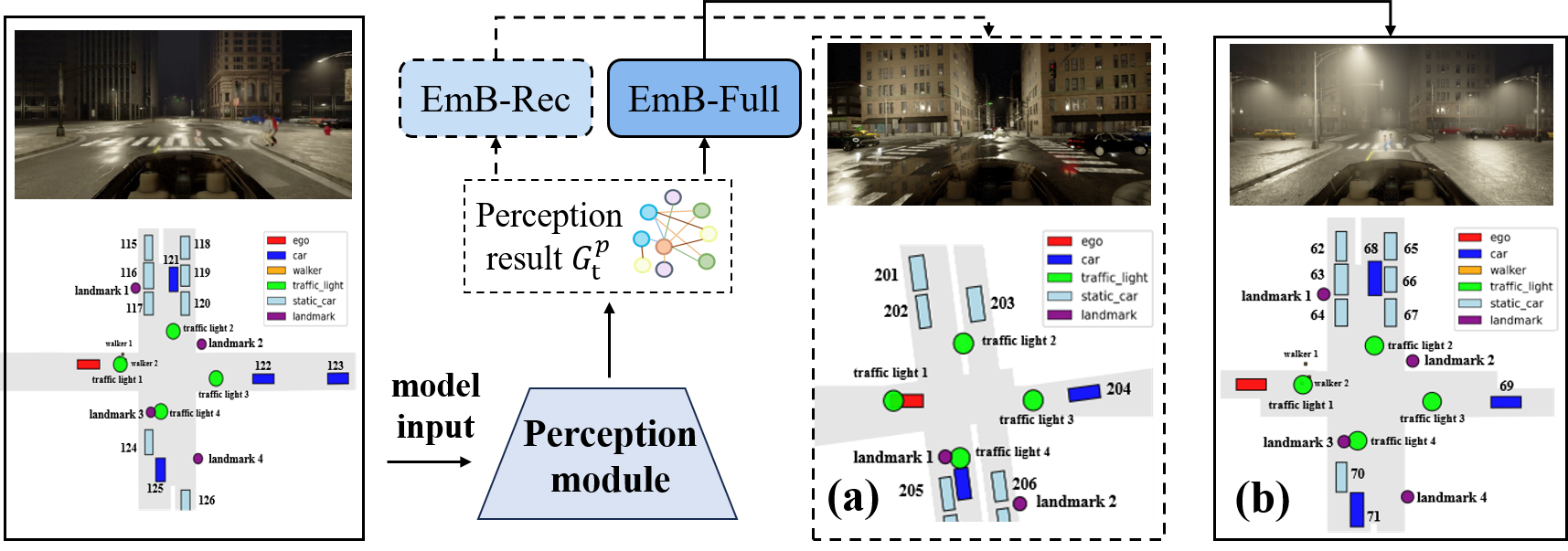}
    \caption{\textbf{Visualization of retrieval quality.} (a) Retrieval scenario from Emb-Rec (b) Retrieval scenario from Emb-Full}
    \label{fig:retrieval_quality}
\end{figure}

\paragraph{Embedding Optimization Objectives.}
Finally, we validate the training objectives of the embedding model by comparing \textit{Emb-Rec} (row 3) with \textit{Emb-Full} (row 4). The \textit{Emb-Rec} variant, trained only with the restoration loss, preserves graph structure but yields inferior performance, indicating that reconstruction alone is insufficient for learning semantically meaningful embeddings. By further introducing the alignment loss, \textit{Emb-Full} produces a more semantically organized embedding space. As shown in Fig. \ref{fig:retrieval_quality}, this leads to retrieved priors with higher semantic consistency, which in turn improves downstream planning performance.

\subsubsection{objective of knowledge embedding}
We analyze the sensitivity of the model to the size of the retrieval vector database $N_D$. As illustrated in Fig. \ref{fig:speed_comp}, enlarging the database generally improves driving performance by providing a broader pool of historical priors. However, this benefit comes at the cost of increased retrieval latency. These results reveal a clear trade-off between planning performance and retrieval efficiency, indicating that the database size should be chosen to balance driving score and inference speed.

We further study whether incorporating the current perception result , $G_t^p$, into the retrieval sequence is necessary. We compare a variant using only retrieved historical priors with our strategy that additionally includes $G_t^p$. As shown in Table \ref{tab:ablation_combined}, incorporating $G_t^p$ brings a clear performance gain. We attribute this improvement to the fact that $G_t^p$ provides an explicit semantic anchor for the current scene, helping the model better align retrieved priors with the current context and suppress irrelevant noise.
\subsubsection{training curriculum}
Finally, we compare the proposed three-stage training strategy with a simplified two-stage baseline in Table \ref{tab:ablation_combined}. In the two-stage setting, the embedding model is first pre-trained, after which the perception module and driving policy are jointly optimized from scratch. In contrast, the three-stage strategy further pre-aligns the perception module before the final end-to-end policy learning stage. The results show that the three-stage design yields better planning performance, demonstrating the effectiveness of staged optimization.

\section{Conclusion}
In this paper, we presented VLADriver-RAG, a holistic retrieval-augmented framework designed to overcome the generalization bottlenecks of parametric VLA models in autonomous driving. By introducing the \textit{Visual-to-Scenario} abstraction and a \textit{Scenario-Aligned Embedding Model}, we shift the retrieval paradigm from inefficient visual matching to precise topological alignment. This structure-aware retrieval enables our query-based VLA backbone to effectively fuse expert priors with real-time perception for robust planning.

\begin{table}[t]
\centering
\caption{Ablation studies on graph prior and training strategy.}
\label{tab:ablation_combined}
\begin{tabular}{lcc|lcc}
\toprule
Graph prior & DS $\uparrow$ & SR $\uparrow$ & Training strategy & DS $\uparrow$ & SR $\uparrow$ \\
\midrule
w/o $G_t^p$  & 88.23 & 69.00 & Two-stage   & 86.76 & 66.95 \\
with $G_t^p$ & 89.12 & 70.42 & Three-stage & 89.12 & 70.42 \\
\bottomrule
\end{tabular}
\end{table}Extensive evaluations on the B2D benchmark confirm that our framework establishes a new SOTA, demonstrating that integrating non-parametric, logic-driven memory is essential for navigating complex open-world environments. Future work will explore scaling the scenario database to millions of entries and investigating continual learning mechanisms to dynamically update the knowledge base from online experiences.

\printbibliography

@inproceedings{chen2024internvl,
  title={Internvl: Scaling up vision foundation models and aligning for generic visual-linguistic tasks},
  author={Chen, Zhe and Wu, Jiannan and Wang, Wenhai and Su, Weijie and Chen, Guo and Xing, Sen and Zhong, Muyan and Zhang, Qinglong and Zhu, Xizhou and Lu, Lewei and others},
  booktitle={Proceedings of the IEEE/CVF conference on computer vision and pattern recognition},
  pages={24185--24198},
  year={2024}
}

@article{wang2024qwen2,
  title={Qwen2-vl: Enhancing vision-language model's perception of the world at any resolution},
  author={Wang, Peng and Bai, Shuai and Tan, Sinan and Wang, Shijie and Fan, Zhihao and Bai, Jinze and Chen, Keqin and Liu, Xuejing and Wang, Jialin and Ge, Wenbin and others},
  journal={arXiv preprint arXiv:2409.12191},
  year={2024}
}

@inproceedings{renz2025simlingo,
  title={Simlingo: Vision-only closed-loop autonomous driving with language-action alignment},
  author={Renz, Katrin and Chen, Long and Arani, Elahe and Sinavski, Oleg},
  booktitle={Proceedings of the Computer Vision and Pattern Recognition Conference},
  pages={11993--12003},
  year={2025}
}

@article{fu2025orion,
  title={Orion: A holistic end-to-end autonomous driving framework by vision-language instructed action generation},
  author={Fu, Haoyu and Zhang, Diankun and Zhao, Zongchuang and Cui, Jianfeng and Liang, Dingkang and Zhang, Chong and Zhang, Dingyuan and Xie, Hongwei and Wang, Bing and Bai, Xiang},
  journal={arXiv preprint arXiv:2503.19755},
  year={2025}
}

@article{black2024pi_0,
  title={{$\pi_0$}: A Vision-Language-Action Flow Model for General Robot Control},
  author={Black, Kevin and Brown, Noah and Driess, Danny and Esmail, Adnan and Equi, Michael and Finn, Chelsea and Fusai, Niccolo and Groom, Lachy and Hausman, Karol and Ichter, Brian and others},
  journal={arXiv preprint arXiv:2410.24164},
  year={2024}
}

@article{intelligence2025pi_,
  title={{$\pi_{0.5}$}: a Vision-Language-Action Model with Open-World Generalization},
  author={Intelligence, Physical and Black, Kevin and Brown, Noah and Darpinian, James and Dhabalia, Karan and Driess, Danny and Esmail, Adnan and Equi, Michael and Finn, Chelsea and Fusai, Niccolo and others},
  journal={arXiv preprint arXiv:2504.16054},
  year={2025}
}

@article{hussien2025rag,
  title={Rag-based explainable prediction of road users behaviors for automated driving using knowledge graphs and large language models},
  author={Hussien, Mohamed Manzour and Melo, Angie Nataly and Ballardini, Augusto Luis and Maldonado, Carlota Salinas and Izquierdo, Rub{\'e}n and Sotelo, Miguel Angel},
  journal={Expert Systems with Applications},
  volume={265},
  pages={125914},
  year={2025},
  publisher={Elsevier}
}

@article{sun2025curriculum,
  title={Curriculum Engineering: Structured Learning for Large Language Models (LLMs) Through Curriculum Based Retrieval},
  author={Sun, Kexin and Zhao, Zhiheng and Yang, Hongxia and Zhang, Jie and Huang, George Q},
  journal={IEEE Transactions on Industrial Informatics},
  year={2025},
  publisher={IEEE}
}

@article{peifeng2024joint,
  title={Joint knowledge graph and large language model for fault diagnosis and its application in aviation assembly},
  author={Peifeng, LIU and Qian, Lu and Zhao, Xingwei and Tao, Bo},
  journal={IEEE Transactions on Industrial Informatics},
  volume={20},
  number={6},
  pages={8160--8169},
  year={2024},
  publisher={IEEE}
}

@article{yuan2024rag,
  title={Rag-driver: Generalisable driving explanations with retrieval-augmented in-context learning in multi-modal large language model},
  author={Yuan, Jianhao and Sun, Shuyang and Omeiza, Daniel and Zhao, Bo and Newman, Paul and Kunze, Lars and Gadd, Matthew},
  journal={arXiv preprint arXiv:2402.10828},
  year={2024}
}

@inproceedings{li2024ego,
  title={Is ego status all you need for open-loop end-to-end autonomous driving?},
  author={Li, Zhiqi and Yu, Zhiding and Lan, Shiyi and Li, Jiahan and Kautz, Jan and Lu, Tong and Alvarez, Jose M},
  booktitle={Proceedings of the IEEE/CVF Conference on Computer Vision and Pattern Recognition},
  pages={14864--14873},
  year={2024}
}

@inproceedings{jia2023think,
  title={Think twice before driving: Towards scalable decoders for end-to-end autonomous driving},
  author={Jia, Xiaosong and Wu, Penghao and Chen, Li and Xie, Jiangwei and He, Conghui and Yan, Junchi and Li, Hongyang},
  booktitle={Proceedings of the IEEE/CVF Conference on Computer Vision and Pattern Recognition},
  pages={21983--21994},
  year={2023}
}

@article{yu2024visrag,
  title={Visrag: Vision-based retrieval-augmented generation on multi-modality documents},
  author={Yu, Shi and Tang, Chaoyue and Xu, Bokai and Cui, Junbo and Ran, Junhao and Yan, Yukun and Liu, Zhenghao and Wang, Shuo and Han, Xu and Liu, Zhiyuan and others},
  journal={arXiv preprint arXiv:2410.10594},
  year={2024}
}

@article{hwang2024emma,
  title={Emma: End-to-end multimodal model for autonomous driving},
  author={Hwang, Jyh-Jing and Xu, Runsheng and Lin, Hubert and Hung, Wei-Chih and Ji, Jingwei and Choi, Kristy and Huang, Di and He, Tong and Covington, Paul and Sapp, Benjamin and others},
  journal={arXiv preprint arXiv:2410.23262},
  year={2024}
}

@article{wang2024omnidrive,
  title={Omnidrive: A holistic llm-agent framework for autonomous driving with 3d perception, reasoning and planning},
  author={Wang, Shihao and Yu, Zhiding and Jiang, Xiaohui and Lan, Shiyi and Shi, Min and Chang, Nadine and Kautz, Jan and Li, Ying and Alvarez, Jose M},
  journal={CoRR},
  year={2024}
}

@inproceedings{xing2025openemma,
  title={Openemma: Open-source multimodal model for end-to-end autonomous driving},
  author={Xing, Shuo and Qian, Chengyuan and Wang, Yuping and Hua, Hongyuan and Tian, Kexin and Zhou, Yang and Tu, Zhengzhong},
  booktitle={Proceedings of the Winter Conference on Applications of Computer Vision},
  pages={1001--1009},
  year={2025}
}

@article{team2023gemini,
  title={Gemini: a family of highly capable multimodal models},
  author={Team, Gemini and Anil, Rohan and Borgeaud, Sebastian and Alayrac, Jean-Baptiste and Yu, Jiahui and Soricut, Radu and Schalkwyk, Johan and Dai, Andrew M and Hauth, Anja and Millican, Katie and others},
  journal={arXiv preprint arXiv:2312.11805},
  year={2023}
}

@article{jiang2024senna,
  title={Senna: Bridging large vision-language models and end-to-end autonomous driving},
  author={Jiang, Bo and Chen, Shaoyu and Liao, Bencheng and Zhang, Xingyu and Yin, Wei and Zhang, Qian and Huang, Chang and Liu, Wenyu and Wang, Xinggang},
  journal={arXiv preprint arXiv:2410.22313},
  year={2024}
}

@article{tian2024drivevlm,
  title={Drivevlm: The convergence of autonomous driving and large vision-language models},
  author={Tian, Xiaoyu and Gu, Junru and Li, Bailin and Liu, Yicheng and Wang, Yang and Zhao, Zhiyong and Zhan, Kun and Jia, Peng and Lang, Xianpeng and Zhao, Hang},
  journal={arXiv preprint arXiv:2402.12289},
  year={2024}
}

@article{wang2023drivemlm,
  title={Drivemlm: Aligning multi-modal large language models with behavioral planning states for autonomous driving},
  author={Wang, Wenhai and Xie, Jiangwei and Hu, ChuanYang and Zou, Haoming and Fan, Jianan and Tong, Wenwen and Wen, Yang and Wu, Silei and Deng, Hanming and Li, Zhiqi and others},
  journal={arXiv preprint arXiv:2312.09245},
  year={2023}
}

@inproceedings{shao2024lmdrive,
  title={Lmdrive: Closed-loop end-to-end driving with large language models},
  author={Shao, Hao and Hu, Yuxuan and Wang, Letian and Song, Guanglu and Waslander, Steven L and Liu, Yu and Li, Hongsheng},
  booktitle={Proceedings of the IEEE/CVF Conference on Computer Vision and Pattern Recognition},
  pages={15120--15130},
  year={2024}
}

@inproceedings{guu2020retrieval,
  title={Retrieval augmented language model pre-training},
  author={Guu, Kelvin and Lee, Kenton and Tung, Zora and Pasupat, Panupong and Chang, Mingwei},
  booktitle={International conference on machine learning},
  pages={3929--3938},
  year={2020},
  organization={PMLR}
}

@article{lewis2020retrieval,
  title={Retrieval-augmented generation for knowledge-intensive nlp tasks},
  author={Lewis, Patrick and Perez, Ethan and Piktus, Aleksandra and Petroni, Fabio and Karpukhin, Vladimir and Goyal, Naman and K{\"u}ttler, Heinrich and Lewis, Mike and Yih, Wen-tau and Rockt{\"a}schel, Tim and others},
  journal={Advances in neural information processing systems},
  volume={33},
  pages={9459--9474},
  year={2020}
}

@inproceedings{borgeaud2022improving,
  title={Improving language models by retrieving from trillions of tokens},
  author={Borgeaud, Sebastian and Mensch, Arthur and Hoffmann, Jordan and Cai, Trevor and Rutherford, Eliza and Millican, Katie and Van Den Driessche, George Bm and Lespiau, Jean-Baptiste and Damoc, Bogdan and Clark, Aidan and others},
  booktitle={International conference on machine learning},
  pages={2206--2240},
  year={2022},
  organization={PMLR}
}

@inproceedings{jiang2023active,
  title={Active retrieval augmented generation},
  author={Jiang, Zhengbao and Xu, Frank F and Gao, Luyu and Sun, Zhiqing and Liu, Qian and Dwivedi-Yu, Jane and Yang, Yiming and Callan, Jamie and Neubig, Graham},
  booktitle={Proceedings of the 2023 Conference on Empirical Methods in Natural Language Processing},
  pages={7969--7992},
  year={2023}
}

@article{asai2024self,
  title={Self-rag: Learning to retrieve, generate, and critique through self-reflection},
  author={Asai, Akari and Wu, Zeqiu and Wang, Yizhong and Sil, Avirup and Hajishirzi, Hannaneh},
  year={2024},
  publisher={ICLR}
}

@article{chang2025driving,
  title={Driving-RAG: Driving Scenarios Embedding, Search, and RAG Applications},
  author={Chang, Cheng and Ge, Jingwei and Guo, Jiazhe and Guo, Zelin and Jiang, Binghong and Li, Li},
  journal={arXiv preprint arXiv:2504.04419},
  year={2025}
}

@article{jia2025spatial,
  title={Spatial Retrieval Augmented Autonomous Driving},
  author={Jia, Xiaosong and Zhang, Chenhe and Jiang, Yule and Wong, Songbur and Zhang, Zhiyuan and Chen, Chen and Zhang, Shaofeng and Zhou, Xuanhe and Yang, Xue and Yan, Junchi and others},
  journal={arXiv preprint arXiv:2512.06865},
  year={2025}
}

@article{chang2024vistascenario,
  title={Vistascenario: Interaction scenario engineering for vehicles with intelligent systems for transport automation},
  author={Chang, Cheng and Zhang, Jiawei and Ge, Jingwei and Zhang, Zuo and Wei, Junqing and Li, Li and Wang, Fei-Yue},
  journal={IEEE Transactions on Intelligent Vehicles},
  year={2024},
  publisher={IEEE}
}

@article{wu2022trajectory,
  title={Trajectory-guided control prediction for end-to-end autonomous driving: A simple yet strong baseline},
  author={Wu, Penghao and Jia, Xiaosong and Chen, Li and Yan, Junchi and Li, Hongyang and Qiao, Yu},
  journal={Advances in Neural Information Processing Systems},
  volume={35},
  pages={6119--6132},
  year={2022}
}

@inproceedings{jia2023driveadapter,
  title={Driveadapter: Breaking the coupling barrier of perception and planning in end-to-end autonomous driving},
  author={Jia, Xiaosong and Gao, Yulu and Chen, Li and Yan, Junchi and Liu, Patrick Langechuan and Li, Hongyang},
  booktitle={Proceedings of the IEEE/CVF International Conference on Computer Vision},
  pages={7953--7963},
  year={2023}
}

@inproceedings{hu2023planning,
  title={Planning-oriented autonomous driving},
  author={Hu, Yihan and Yang, Jiazhi and Chen, Li and Li, Keyu and Sima, Chonghao and Zhu, Xizhou and Chai, Siqi and Du, Senyao and Lin, Tianwei and Wang, Wenhai and others},
  booktitle={Proceedings of the IEEE/CVF conference on computer vision and pattern recognition},
  pages={17853--17862},
  year={2023}
}

@inproceedings{jiang2023vad,
  title={Vad: Vectorized scene representation for efficient autonomous driving},
  author={Jiang, Bo and Chen, Shaoyu and Xu, Qing and Liao, Bencheng and Chen, Jiajie and Zhou, Helong and Zhang, Qian and Liu, Wenyu and Huang, Chang and Wang, Xinggang},
  booktitle={Proceedings of the IEEE/CVF International Conference on Computer Vision},
  pages={8340--8350},
  year={2023}
}

@article{zhai2023rethinking,
  title={Rethinking the open-loop evaluation of end-to-end autonomous driving in nuscenes},
  author={Zhai, Jiang-Tian and Feng, Ze and Du, Jinhao and Mao, Yongqiang and Liu, Jiang-Jiang and Tan, Zichang and Zhang, Yifu and Ye, Xiaoqing and Wang, Jingdong},
  journal={arXiv preprint arXiv:2305.10430},
  year={2023}
}

@inproceedings{sima2024drivelm,
  title={Drivelm: Driving with graph visual question answering},
  author={Sima, Chonghao and Renz, Katrin and Chitta, Kashyap and Chen, Li and Zhang, Hanxue and Xie, Chengen and Bei{\ss}wenger, Jens and Luo, Ping and Geiger, Andreas and Li, Hongyang},
  booktitle={European conference on computer vision},
  pages={256--274},
  year={2024},
  organization={Springer}
}

@article{xu2024drivegpt4,
  title={Drivegpt4: Interpretable end-to-end autonomous driving via large language model},
  author={Xu, Zhenhua and Zhang, Yujia and Xie, Enze and Zhao, Zhen and Guo, Yong and Wong, Kwan-Yee K and Li, Zhenguo and Zhao, Hengshuang},
  journal={IEEE Robotics and Automation Letters},
  volume={9},
  number={10},
  pages={8186--8193},
  year={2024},
  publisher={IEEE}
}

\end{document}